\documentclass[nohyperref]{article}

\usepackage{microtype}
\usepackage{graphicx}
\usepackage{subfigure}
\usepackage{bbm}
\usepackage[frozencache,cachedir=.]{minted}

\usepackage{booktabs} % for professional tables
\usepackage{hyperref}

\usepackage[accepted]{icml2022}

\usepackage{amsmath}
\usepackage{amssymb}
\usepackage{mathtools}
\usepackage{amsthm}
\usepackage{thm-restate}
\usepackage[capitalize,noabbrev]{cleveref}

\theoremstyle{plain}
\newtheorem{theorem}{Theorem}[section]
\newtheorem{proposition}[theorem]{Proposition}
\newtheorem{lemma}[theorem]{Lemma}

\theoremstyle{definition}

\theoremstyle{remark}

\usepackage[textsize=tiny]{todonotes}
\usepackage{xcolor}
\usepackage{colortbl}
\usepackage{xspace}

\definecolor{ETHBlue}{RGB}{33,92,175}   % blue
\definecolor{ETHGreen}{RGB}{98,115,19}      % green
\definecolor{ETHPurpleDark}{RGB}{140,10,89} % purple
\definecolor{ETHPurple}{RGB}{163,7,116} % purple
\definecolor{ETHGray}{RGB}{111,111,111} % gray
\definecolor{ETHRed}{RGB}{183,53,45}    % red
\definecolor{ETHPetrol}{RGB}{0,120,148} % green/blue
\definecolor{ETHBronze}{RGB}{142,103,19}    % bronze
\definecolor{ETHOrange}{RGB}{230, 100, 50}

\colorlet{MacroColor}{black}
\colorlet{MACROCOLOR}{black}

\newcommand{\mymacro}[1]{{\color{MacroColor} #1}}

\newcommand{\param}{\mymacro{{\boldsymbol{\theta}}}}

\newcommand{\params}{\mymacro{{\boldsymbol{\Theta}}}}

\newcommand{\vv}{\mymacro{\boldsymbol{v}}}

\newcommand{\ee}[1]{\mymacro{\boldsymbol{e}_{#1}}}
\newcommand{\range}{\mymacro{{\small \textsf{range}}}}

\newcommand{\fantope}{\mathcal{F}_\targetdimension}
\newcommand{\fantopegrey}{\mathcal{F}_K}

\newcommand{\tr}{\mymacro{\mathrm{tr}}}
\newcommand{\symmetricmatrices}{\mymacro{\mathcal{S}^D}}

\DeclareMathOperator*{\argmin}{\mymacro{\mathrm{argmin}}}

\newcommand{\bertadv}{\textsf{BERT-adv}\xspace}
\newcommand{\bertafinetuned}{\textsf{BERT-finetuned}\xspace}
\newcommand{\bertafreezed}{\textsf{BERT-frozen}\xspace}
\newcommand{\CLS}{\textsf{[CLS]}\xspace}

\newcommand{\method}{RLACE\xspace}
\newcommand{\defeq}[0]{\mathrel{\stackrel{\textnormal{\tiny def}}{=}}}
\usepackage{inconsolata}

\usepackage[capitalize,noabbrev]{cleveref}
\crefname{section}{\S}{\S\S}
\Crefname{section}{\S}{\S\S}
\crefname{table}{Tab.}{}
\crefname{figure}{Fig.}{}
\crefname{algorithm}{Alg.}{}
\crefname{appendix}{App.}{}
\crefname{lemma}{Lemma}{}
\Crefname{theorem}{Theorem}{}
\crefname{prop}{Proposition}{}
\crefname{cor}{Corollary}{}
\crefname{myexample}{Example}{}
\crefname{align}{}{}
\crefname{equation}{Eq.}{Eqs.}

\newcommand{\linkfun}{\mymacro{g}}
\newcommand{\R}{\mymacro{\mathbb{R}}}
\renewcommand{\AA}{\mymacro{\boldsymbol{A}}}
\newcommand{\VV}{\mymacro{\boldsymbol{V}}}
\newcommand{\LLambda}{\mymacro{\boldsymbol{\Lambda}}}
\newcommand{\XX}{\mymacro{\boldsymbol{X}}}
\newcommand{\PP}{\mymacro{\boldsymbol{P}}}
\newcommand{\PPgrey}{\textcolor{gray}{{\boldsymbol{P}}}}

\newcommand{\PPzero}{\mymacro{\boldsymbol{P}_0}}

\newcommand{\eigenvalue}[1]{\mymacro{\lambda_{#1}}}
\newcommand{\eigenvector}[1]{{\mymacro{\vv_{#1}}}}

\newcommand{\targetdimension}{\mymacro{K}}
\newcommand{\BB}{\mymacro{\boldsymbol{B}}}
\newcommand{\WW}{\mymacro{\boldsymbol{W}}}
\newcommand{\ID}{\mymacro{\boldsymbol{I}_D}}
\newcommand{\Ik}{\mymacro{\boldsymbol{I}_{\targetdimension}}}

\newcommand{\word}[1]{{\mymacro{``#1''}}}
\newcommand{\concept}[1]{\mymacro{\textsc{#1}}}

\newcommand{\xx}[1]{\mymacro{\boldsymbol{x}_{#1}}}

\newcommand{\labely}[1]{\mymacro{y}_{#1}}
\newcommand{\labelyhat}[1]{\mymacro{\widehat{y}_{#1}}}
\newcommand{\labelset}{\mymacro{\mathcal{Y}}}
\newcommand{\yy}{\mymacro{\boldsymbol{y}}}

\newcommand{\inv}[1]{#1^{-1}}
\newcommand{\link}[1]{\mymacro{\inv{g}\!\left(#1\right)}}
\newcommand{\rem}[1]{\mymacro{r(#1)}}

\newcommand{\defn}[1]{\mymacro{\textbf{#1}}}

\newcommand{\projBComp}{{ {\footnotesize \textsf{proj}}_{\BB_{\perp}}}}
\newcommand{\lanullspace}{{ {\footnotesize \textsf{null}}}}
\newcommand{\larange}{{ {\footnotesize \textsf{range}}}}

\newcommand{\valpha}{\boldsymbol{\valpha}}
\newcommand{\projk}{\mymacro{\mathcal{P}_{\targetdimension}}}

\newcommand{\loss}{\mymacro{\ell}}

\newcommand*{\nameadjunct}{\relax}
\makeatletter
\renewcommand*{\NAT@nmfmt}[1]{\NAT@up #1\nameadjunct}
\makeatother

\newcommand*{\citeposs}[2][]{%
  \begingroup
  \renewcommand*{\nameadjunct}{'s}%
  \citet[#1]{#2}%
  \endgroup
}

\newcommand{\dataset}{{\mathcal{D}}}

    \theoremstyle{plain}
    \newtheoremstyle{TheoremNum}
        {\topsep}{\topsep}              %%% space between body and thm
        {\itshape}                      %%% Thm body font
        {}                              %%% Indent amount (empty = no indent)
        {\bfseries}                     %%% Thm head font
        {.}                             %%% Punctuation after thm head
        { }                             %%% Space after thm head
        {\thmname{#1}\thmnote{ \bfseries #3}}%%% Thm head spec
    \theoremstyle{TheoremNum}

\newtheorem{myexample}{Example}

\icmltitlerunning{Linear Adversarial Concept Erasure}

\begin{document}

\twocolumn[
\icmltitle{Linear Adversarial Concept Erasure}

\icmlsetsymbol{equal}{*}

\begin{icmlauthorlist}
\icmlauthor{Shauli Ravfogel}{biu,ai2}
\icmlauthor{Michael Twiton}{ind}
\icmlauthor{Yoav Goldberg}{biu,ai2}
\icmlauthor{Ryan Cotterell}{eth}
\end{icmlauthorlist}

\icmlaffiliation{biu}{Department of Computer Science, Bar Ilan University}
\icmlaffiliation{ai2}{Allen Institute for Artificial Intelligence}
\icmlaffiliation{eth}{ETH Zürich}
\icmlaffiliation{ind}{Independent researcher}

\icmlcorrespondingauthor{Shauli Ravfogel}{shauli.ravfogel@gmail.com}
\icmlcorrespondingauthor{Michael Twiton}{mtwito101@gmail.com}
\icmlcorrespondingauthor{Yoav Goldberg}{yoav.goldberg@gmail.com}
\icmlcorrespondingauthor{Ryan Cotterell}{ryan.cotterell@inf.ethz.ch}

\icmlkeywords{Machine Learning, intervention, concept, information-removal, interpretation, bias, ICML}

\vskip 0.3in
]



\printAffiliationsAndNotice{} % otherwise use the standard text.

\begin{abstract}
Modern neural models trained on textual data rely on pre-trained representations that emerge without direct supervision. As these representations are increasingly being used in real-world applications, the inability to \emph{control} their content becomes an increasingly important problem. 
We formulate the problem of identifying and erasing a linear subspace that corresponds to a given concept, in order to prevent linear predictors from recovering the concept. We model this problem as a constrained, linear maximin game, and show that existing solutions are generally not optimal for this task. We derive a closed-form solution for certain objectives, and propose a convex relaxation, \method, that works well for others. When evaluated in the context of binary gender removal, the method recovers a low-dimensional subspace whose removal mitigates bias by intrinsic and extrinsic evaluation. We show that the method is highly expressive, effectively mitigating bias in deep nonlinear classifiers while maintaining tractability and interpretability.
\newline
\newline
\vspace{1.5em} 
\hspace{.5em}\includegraphics[width=1.25em,height=1.25em]{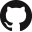}\hspace{.75em}\parbox{\dimexpr\linewidth-4\fboxsep-2\fboxrule}{\url{https://github.com/shauli-ravfogel/rlace-icml}}
\end{abstract}
\vspace{-40pt}

\begin{figure}[]
    \centering
    \includegraphics[width=0.96\columnwidth]{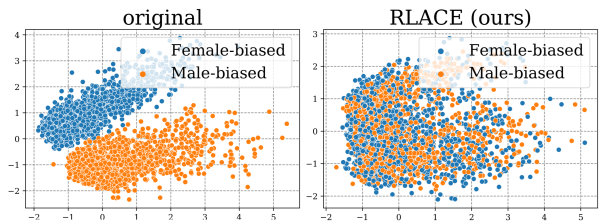}
    \caption{Removal of gender information from GloVe representations using \method, after PCA (Experiment \cref{sec:embeddings}). Left: original space; Right: after a rank-1 \method projection. Word vectors are colored according to their being male-biased or female-biased.}
    \label{fig:pca}
\vspace{-20pt}
\end{figure}

\section{Introduction}
We are interested in the question of removing information from a given real-valued vector representation, e.g., representations that are obtained via neural language encoders of text \cite{melamud-etal-2016-context2vec, peters-etal-2018-deep, howard-ruder-2018-universal, devlin-etal-2019-bert}.
Specifically, we ask the following: Given a set of vectors $\xx{1}, \ldots, \xx{N} \in \R^D$ and a labeling $\labely{1},\ldots,\labely{N}$, with each $\labely{n} \in \labelset$,\footnote{Throughout this paper, we take $\labelset = \{0, 1\}$.} a label set of concepts, can we derive a function $\rem{\cdot}$ such that the resulting vectors $\rem{\xx{1}}, \ldots, \rem{\xx{N}}$ are \emph{not} predictive of the concept labels $\labely{1}, \ldots ,\labely{N}$, but $\rem{\xx{n}}$ preserves the information found in $\xx{n}$ as much as possible?
However, unlike methods \citep{edwards2015censoring, chen2018adversarial, xie2017controllable,elazar-goldberg-2018-adversarial, zhang2018mitigating} that require a modification to the training process, here we are interested in \textit{post-hoc} methods, which assume a fixed, pre-trained encoder (such as GloVe \cite{DBLP:conf/emnlp/PenningtonSM14}, BERT \cite{devlin-etal-2019-bert}, or GPT-2 \cite{radford2019language} and aim to learn an additional function $\rem{\cdot}$ that removes information from the fixed representations.
Thus, this problem generalizes bias mitigation, e.g., by removing the concept of gender, from word representations \cite{bolukbasi2016man}.\looseness=-1

In this article, we focus on the special case where the function $\rem{\cdot}$ is a linear transformation---specifically, we aim to identify and remove a linear concept subspace from the representation using an orthogonal projection matrix in such a manner that prevents any linear classifier from recovering the value of the concept. 
By restricting ourselves to the linear case, we obtain a practical solution while also enjoying the increased interpretability of linear methods. 
However, by imposing the constraint that the linear transformation is a non-low-rank orthogonal projection matrix, we enforce that the linear transformation is minimally invasive. 

Isolating a linear concept space in representations of text was pioneered by \citet{bolukbasi2016man}, who used principal component analysis to identify a linear gender bias subspace in static word representations. 
After identifying the bias subspace, \citet{bolukbasi2016man} gave a recipe for mitigating gender bias from the word representations. 
However, \citet{gonen-goldberg-2019-lipstick} later demonstrated that \citeposs{bolukbasi2016man} method does not exhaustively remove gender bias.
Indeed, a linear classifier trained on the modified representations can still recover the gender labels initially associated with each representation.
In an attempt to improve upon \citet{bolukbasi2016man}, \citep[INLP;][]{ravfogel-etal-2020-null} introduced iterative nullspace projection (INLP), which estimates a linear gender subspace by first training a classifier to predict gender, and then projecting onto the nullspace of learned classifier's weights; \citet{ravfogel-etal-2020-null} found their method competitive.\looseness=-1

We provide a thorough analysis of the problem of identifying and neutralizing linear concept subspaces formalized as a maximin game \citep{von1947theory}.
We contend our formalization of linear adversarial concept removal offers us the best of two worlds.
On one hand, in some cases, we can maintain the superior performance often witnessed in the adversarial paradigm.
On the other, we maintain a more interpretable concept space due to our linearity assumption.
In several cases, such as linear regression and Rayleigh quotient maximization, we are able to derive a closed-form solution to the maximin problem.
For the case of classification loss, e.g., logistic regression, we develop a convex relaxation, \textbf{Relaxed Linear Adversarial Concept Erasure (\method)}, that allows us to find a good solution in practice. 
For concreteness, in our experiments, we follow the motivating example of removing information predictive of binary gender, and find the method effective in mitigating bias in both contextualized and static representations.\looseness=-1

\section{Linear Maximin Games}
\label{sec:formulation}
This section focuses on the mathematical preliminaries necessary to develop linear adversarial concept removal, a formulation that allows us to impose a structure on the adversarial intervention.
Specifically, we formulate the problem as a maximin game between a predictor that aims to predict a quantity that operationalizes the concept (e.g., binary gender) and an adversary that tries to hinder the prediction by projecting the representations onto a subspace of predefined dimensionality. 
By constraining the adversarial intervention to a linear projection, we maintain the advantages of linear methods---interpretability and transparency---while directly optimizing an expressive objective that aims to prevent \textit{any} linear model from predicting the concept of interest.

\subsection{Notation and Generalized Linear Modeling}
We overview generalized linear modeling \cite{nelder1972generalized} as a framework for concept erasure.\looseness=-1

\paragraph{Notation.} 
Assume we are given a dataset $\dataset = \{(\xx{n}, \labely{n})\}_{n=1}^N$ of $N$ response--representation pairs, where the response variables $\labely{n}$ represent the information to be neutralized, e.g., binary gender.
In this article, we take $\labely{n} \in \labelset \defeq \{0, 1\}$ and $\xx{n} \in \R^D$ to be a $D$-dimensional real column vector.\footnote{We could have also formulated the problem where $\labely{n}$ was also a vector. 
We have omitted this generalization for simplicity.}
We use the notation $\XX=[\xx{1}; \cdots; \xx{N}]^{\top} \in \R^{N \times D}$ to denote a matrix containing the inputs, and $\yy = [\labely{1}, \ldots, \labely{n}]^{\top} \in \R^N$ to denote a column vector containing all the dependent variables.

\paragraph{Generalized Linear Models.} 
Let $\params \subset \R^D$ be a compact set of parameters, and let $\param \in \params$ be a real column vector of parameters. 
A generalized linear model consists of a linear predictor of the form $\param^{\top} \xx{n}$ coupled with a \defn{link function} $\linkfun(\cdot)$, which allows us to relate a linear prediction to the response in a more nuanced (perhaps non-linear) way. 
Denoting the link function's inverse as $\link{\cdot}$, we write the prediction of a generalized linear model as $\labelyhat{n} \defeq \link{\param^{\top} \xx{n}}$.
We additionally assume a \defn{loss} function $\loss(\labely{n}, \labelyhat{n}) \geq 0$, a non-negative function of the true response $\labely{n}$ and a predicted response $\labelyhat{n}$, that is to be minimized. 
By changing the link and loss functions, we obtain different problems such as linear regression, Rayleigh quotient minimization, logistic regression classification, and others.
Using the above notation, this paper considers the generalized linear classification objective\looseness=-1
\begin{equation}
    \sum_{n=1}^N \loss\left(\labely{n}, \labelyhat{n} \right) = \sum_{n=1}^N \loss\left(\labely{n}, \link{\param^{\top} \xx{n}} \right). \label{eq:glm}
\end{equation}
We seek to minimize \cref{eq:glm} with respect to $\param \in \params$ in order to learn a good predictor of the concept labels.\looseness=-1

\subsection{The Linear Bias Subspace Hypothesis}
Consider a collection $\{\xx{m}\}_{m=1}^M$ of $M$ representations where each $\xx{m} \in \R^D$.
The linear bias subspace hypothesis \cite{bolukbasi2016man,vargas} posits that there exists a linear subspace $\BB \subseteq \R^D$ that (fully) contains gender bias information within representations $\{\xx{m}\}_{m=1}^M$.\footnote{While \citet{bolukbasi2016man} and \citet{vargas} focused on bias mitigation, their notion of a bias subspace can be extended to \emph{any} concept, and we do so here.}
It follows from this hypothesis that one strategy for the removal of gender information from representations is to i) identify the subspace $\BB$, and ii) project the representations onto the orthogonal complement of $\BB$, i.e., re-define every representation $\xx{m}$ in our collection as $\projBComp(\xx{m})$.
Basic linear algebra tells us that the operation $\projBComp$ may be represented by an \defn{orthogonal projection matrix}, i.e., a symmetric matrix $\PP$ such that $\PP^2=\PP$ and $\projBComp(\xx{m})=\PP\xx{m}$.
In this formulation, we have that $\lanullspace(\PP)$ is the linear subspace $\BB$ that encodes the bias, and $\larange(\PP)$ is its orthogonal complement, i.e., the non-bias subspace.
Intuitively, an orthogonal projection matrix onto a subspace maps a vector to its closest neighbor in the subspace. In our case, the projection maps a vector to the closest vector in the subspace that excludes the bias subspace.

\subsection{Linear Maximin Games}
We are now in a position to define a linear maximin game that adversarially identifies and removes a linear bias subspace. 
Following \citet{ravfogel-etal-2020-null}, we search for an orthogonal projection matrix $\PP$ that projects onto $\BB_{\perp}$, i.e., the orthogonal complement of the bias subspace $\BB$.
We define $\projk$ as the set of all $D \times D$ orthogonal projection matrices of rank $D-\targetdimension$.
More formally, we have that $\PP \in \projk \Leftrightarrow \PP=\ID - \WW^{\top}\WW, \WW \in \R^{\targetdimension \times D}, \WW\WW^{\top} = \Ik$, where $\ID$ denotes the $D \times D$ identity matrix and $\Ik$ denotes the $\targetdimension \times \targetdimension$ identity matrix. 
The matrix $\PP$'s kernel is the $\targetdimension$-dimensional subspace $\BB=\larange(\WW^{\top}\WW)$.

We now define the following sequential maximin game:\footnote{\color{purple} \textbf{Correction:}
An earlier version of this paper erroneously claimed that \Cref{eq:main-game} is a convex--concave game \cite{kneser1952theoreme, tuy2004minimax}.
The game is actually convex--convex. 
We thank David Schneider-Joseph for pointing out this mistake.  
If the game given in \Cref{eq:relaxed-game} had been convex--concave, the order of the $\min$ and $\max$ would not have been relevant.
However, given that it is \emph{not},  we have adjusted the formulation such that $\max$ precedes the $\min$, which differs from the earlier version of this paper, but has the semantics we originally intended.}
\begin{equation}
    \max_{{\PP \in \projk}}\,\min_{\param \in \params}\,\sum_{n=1}^N \loss\Big(\labely{n},\, \link{\param^{\top} \PP\xx{n}}\Big), \label{eq:main-game}
\end{equation}
where $\targetdimension$, the dimensionality of the bias subspace, is a hyperparameter.
Recall that $\projk$ is the set of all $D \times D$ orthogonal projection matrices of rank $D-\targetdimension$.
We say that pair $(\param^\star, \PP^\star)$ is a solution to  \Cref{eq:main-game} if
\begin{equation}
\begin{aligned}\label{eq:solution}
    \max_{{\PP \in \projk}}\,&\min_{\param \in \params}\,\sum_{n=1}^N \loss\Big(\labely{n},\, \link{\param^{\top} \PP\xx{n}}\Big) \\
    &\quad = \sum_{n=1}^N \loss\Big(\labely{n},\, \link{{\param^\star}^{\top} \PP^\star\xx{n}}\Big).
\end{aligned}
\end{equation}
\Cref{eq:solution} can be thought of as a sequential game where the first player chooses an orthogonal matrix $\PP$ and the second player chooses a parameter vector $\param \in \params$ with knowledge of of the orthogonal matrix $\PP$.
Such a solution to the sequential game, due to the assumed compactness of $\params$ and the compactness of $\projk$, always exists in our setting.
However, it is NP-hard to solve in general \citep{minimaxnphard}.
\cref{eq:main-game} is a special case of the general adversarial training algorithm \cite{DBLP:journals/corr/GoodfellowPMXWOCB14}, but where the adversary is constrained to interact with the input only via an orthogonal projection matrix of rank at most $D-\targetdimension$. 
This constraint enables us to derive principled solutions, while \emph{minimally} changing the input.\footnote{Note that an orthogonal projection of a point onto a subspace gives the \emph{closest} point on that subspace in terms of $L_2$ distance.}\looseness=-1

We now spell out several instantiations of common generalized linear models within the framework of adversarial generalized linear modeling: (i) linear regression, (ii) partial least squares regression, and (iii) logistic regression.
\begin{myexample}[(Linear Regression)]\label{ex:lin}
Consider the loss function $\loss(\labely{n}, \labelyhat{n}) = {||\labely{n} - \labelyhat{n}||}^2$,
and the inverse link function $\link{z} = z$.
Then \cref{eq:main-game} corresponds to
\begin{equation}
\begin{aligned}
   \max_{\PP \in \projk}\,&\min_{\param \in \params}\, \sum_{n=1}^N {||\labely{n} - \param^{\top}\PP \xx{n} ||}^2 \\
   &=  \max_{\PP \in \projk}\,\min_{\param \in \params} \,||\yy - \XX\PP \param ||^2_{\mathrm{F}}.
   \label{exm:regression}
\end{aligned}
\end{equation}
\end{myexample}

\begin{myexample}[(Partial Least Squares Regression)]\label{ex:pls}
Consider the loss function $\loss(\labely{n}, \labelyhat{n}) = (\labely{n} \labelyhat{n})^2$ and inverse link function $\link{z} = z$.
Then \cref{eq:main-game} corresponds to
\begin{equation}
\begin{aligned}
  \max_{\PP \in \projk}\, \min_{\substack{\param \in \params, \\ ||\PP\param||^2 = 1}}\,&\sum_{n=1}^N || \param^{\top} \PP \xx{n} \labely{n}||^2 \\
   &= \max_{\PP \in \projk}\, \min_{\substack{\param \in \params, \\ ||\PP\param||^2 = 1}}\, || \param^{\top} \PP \XX^{\top} \yy||^2_{\mathrm{F}}.
   \label{exm:pls}
\end{aligned}
\end{equation}
where we have additionally placed a constraint on the parameter space on $\param$ post-multiplied by the projection matrix $\PP$, which is, strictly speaking, not part of the formalism.
This means that partial least squares, strictly speaking, does not fit into our paradigm of general linear modeling. 
\end{myexample}

\begin{myexample}[(Logistic Regression)]\label{ex:log}
Consider the loss function $\ell(\labely{n}, \labelyhat{n}) = - \labely{n} \log \labelyhat{n} - (1- \labely{n}) \log (1 - \labelyhat{n})$, and the inverse link function $\link{z} = \frac{\exp z}{1 + \exp z}$.
Then \cref{eq:main-game} corresponds to\looseness=-1
\begin{align}
  &\max_{\PP \in \projk}\,\min_{\param \in \params}  - \Bigg( \sum_{n=1}^N \labely{n} \log \frac{\exp(\param^{\top} \PP \xx{n})} {1 + \exp(\param^{\top}\PP \xx{n})}\nonumber \\
   &\qquad  + (1-\labely{n} ) \log \frac{1} {1 + \exp(\param^{\top}\PP \xx{n})}  \Bigg). \label{exp:logistic-regression}
\end{align}
\end{myexample}

\section{Solving the Linear Maximin Game}
\label{sec:optimality}
At the technical level, this paper asks a simple question: For which pairs of $\loss(\cdot, \cdot)$ and $\link{\cdot}$ can we solve the objective given in \cref{sec:formulation}?
We find a series of satisfying answers.
In the case of linear regression (\cref{ex:lin}) and Rayleigh quotient optimization, e.g., partial least squares regression (\cref{ex:pls}), we derive a closed-form solution.
And, in the case of logistic regression (\cref{ex:log}), we derive a convex relaxation that allows us to solve it efficiently in practice with a gradient-based optimization method.\looseness=-1

\subsection{Linear Regression}
\label{sec:regression}
We begin with the case of linear regression (\cref{ex:lin}).
We show that there exists an optimal solution to \cref{exm:regression} in the following proposition.\looseness=-1

\begin{restatable}{proposition}{regression}
For $\targetdimension=1$, the maximin game given in \Cref{exm:regression} has a solution at $(\PP^\star, \boldsymbol{0})$ where 
\begin{equation}\label{eq:linear-regression-equilibrium}
\PP^\star= \ID - \frac{\XX^{\top}\yy \yy^{\top} \XX}{\yy^{\top}\XX \XX ^{\top} \yy}.
\end{equation}
At this point, the objective evaluates to $||\yy||^2$.\looseness=-1
\label{prop:lin}

\end{restatable}
\begin{proof}
See \Cref{app:regression} for a proof.
\end{proof}
Note that $\PP^\star= \ID - \frac{\XX^{\top}\yy \yy^{\top} \XX}{\yy^{\top}\XX \XX ^{\top} \yy}$ is an orthogonal projection matrix whose null space spans the direction $\XX^{\top} \yy$, the covariance between the representations and responses.
Moreover, because linear regression aims to explain covariance, it suffices to consider a one-dimensional bias subspace.\looseness=-1

\subsection{Partial Least Squares}
\label{sec:Rayleigh}
We now turn to partial least squares regression \cite{wold1973nonlinear} as a representative of a special class of objectives based on the Rayleigh quotient.
The loss function described in \cref{ex:pls} is \emph{not} convex due to the constraint that the parameters have unit norm. 
However, we can still efficiently minimize making use of basic results in linear algebra \cite{horn2012matrix}
Other techniques in this framework include principal component analysis \cite{pearson1901liii} and canonical correlation analysis \cite{hotelling1936rank}.
Recall that we omit the bias term from consideration when analyzing partial least squares. 

We now state a lemma about maximin games in the form of a Rayleigh quotient. 
This lemma allows us to show that \cref{ex:pls} can be solved exactly with an eigendecomposition.\looseness=-1

\begin{restatable}{lemma}{rayleigh}\label{lemma:rayleigh}
Let $\AA \in \R^{D \times D}$ be a symmetric matrix, and let $\AA = \VV^{\top} \LLambda \VV$ be its eigendecomposition.
We order $\AA$'s orthonormal eigenbasis $\{\eigenvector{1}, \ldots, \eigenvector{D}\}$ in an ascending fashion according to the eigenvalues $\eigenvalue{1} \leq \eigenvalue{2} \cdots \leq \eigenvalue{D}$. 
Then, the maximin game
\begin{equation}\label{eq:saddle-point}
    \max_{\PP \in \projk}  \min_{\param \in \params} \frac{\param^{\top} \PP^{\top} \AA \PP \param}{||\PP \param||^2},
\end{equation}
has the solution
\begin{subequations}
\begin{align}
   \PP^\star &= \ID - \sum_{d=1}^{\targetdimension} \eigenvector{d} \eigenvector{d}^{\top} \label{pls-solution2} \\
    \param^\star &= \eigenvector{\targetdimension+1} \label{pls-solution1}.
\end{align}
\end{subequations}
At this point, the objective \cref{eq:saddle-point} evaluates to $\eigenvalue{\targetdimension+1}$. 
\end{restatable}
\begin{proof}
The proof is provided in \cref{app:rayleigh}.
\end{proof}

\begin{proposition}\label{prop:pls}
The partial least squares objective, given in \cref{exm:pls}, is a special case of \Cref{eq:saddle-point} where we define
\begin{equation}
    \AA=\XX^{\top}\yy\yy^{\top}\XX.
\end{equation}
Thus, its solution is given by (\cref{pls-solution1}, \cref{pls-solution2}).
\end{proposition}
\begin{proof}
The adversarial partial least squares objective \cref{ex:pls} is scale invariant. 
It can be equivalently expressed as\looseness=-1
\begin{subequations}
\begin{align}
 \max_{\PP \in \projk}& \min_{\substack{\param \in \params, \\ ||\PP\param||^2 = 1}}\sum_{n=1}^N || \param^{\top} \PP \xx{n} \labely{n}||^2 \\
   &= \max_{\PP \in \projk} \min_{\substack{\param \in \params, \\||\PP\param||^2 = 1}} \,\, \param^{\top}\PP\XX^{\top}\yy\yy^{\top}\XX\PP\param \\
   &=\max_{\PP \in \projk} \min_{\param \in \params}  \frac{\param^{\top}\PP\XX^{\top}\yy\yy^{\top}\XX\PP\param}{||\PP\param||^2}.
\end{align}
\end{subequations}
Now, define $\AA=\XX^{\top}\yy\yy^{\top}\XX$ and the result follows.
\end{proof}

\subsection{Logistic Regression}\label{sec:classification}
We now turn to the most practical setting where we consider logistic regression.
In this case, we propose a practical convex relaxation of the problem.
Note that while our exposition focuses on logistic regression, any other convex loss, e.g., hinge loss, may be substituted in.

We now describe \textbf{Relaxed Linear Adversarial Concept Erasure (\method)}, an effective method to solve the objective \cref{eq:main-game} for classification problems.
To overcome the need to search over all orthogonal projection matrices, we propose to relax $\projk$ to its
\textbf{convex hull}.
In the case of a rank-constrained orthogonal projection matrix, the convex hull is called the \textbf{Fantope} \cite{DBLP:books/cu/BV2014}:
\begin{equation}
\fantope = \{\AA \in \symmetricmatrices \mid \ID \succcurlyeq \AA \succcurlyeq 0, \tr(\AA) = \targetdimension \},
\label{def:fantope}
\end{equation}
where $\symmetricmatrices$ is the of all $D\times D$ real symmetric matrices, $\tr$ is the trace operator, and $\succcurlyeq$ refers to the eigenvalues of the matrix $\AA$.
This yields the following relaxation of \cref{eq:main-game}:
\begin{equation} 
  \max_{{\textcolor{gray}{\PPgrey \in \fantopegrey}}}\, \min_{\param \in \R^D}\,\sum_{n=1}^N \loss(\labely{n},\, \link{\param^{\top}\PP \xx{n}}), \label{eq:relaxed-game}
\end{equation}
where the relaxation is highlighted in grey.

\begin{listing}
\centering
%\pi_{\cotokenizer \circ \modeltokens}$
%\begin{adjustbox}{width=\columnwidth}
\begin{minted}[escapeinside=@@]{python}

def @$\textsc{INLP}$@(@$\{(\xx{n}, \labely{n})\}_{n=1}^N$@, @$\targetdimension$@, @$\loss(\cdot, \cdot)$@, @$\link{\cdot}$@):
  # param: @$\{(\xx{n}, \labely{n})\}_{n=1}^N$@ training data
  # param: @$K$@ bias subspace dimension
  # param: @$\loss(\cdot, \cdot)$@ loss function
  # param: @$\link{\cdot}$@ inverse link function
  @$\PP_0$@ = @$\ID$@
  for @$k$@ in range(@$\targetdimension$@):
    # ties are broken arbitrarily
    @$\param^\star_k$@ = @$\argmin_{\param \in \params} \sum_{n=1}^N \loss\left(\labely{n}, \link{\param^{\top}\PP_{k-1} \xx{n} }\right)$@
    @$\PP_{k}$@ = @$\PP_{k} - \frac{\param^\star_{k}{\param^\star_{k}}^{\top}}{{\param^\star_{k}}^{\top}\param^\star_{k}}$@
    return @$\PP_{\targetdimension}$@
\end{minted}
%\end{adjustbox}
\caption{A Python-esque implementation of INLP.}\label{code:inlp}
\end{listing}

We solve the relaxed objective \cref{eq:relaxed-game} with alternate minimization and maximization over $\param$ and $\PP$, respectively.\footnote{
This procedure is a form of gradient descent--ascent and, thus, is not guaranteed to converge.
However, given promising empirical results, we did not explore more complex algorithms that come with a convergence guarantee.
}
Concretely, we alternate between: (i) holding $\PP$ fixed and taking an unconstrained gradient step over $\param$ towards minimizing the objective, (ii) holding $\param$ fixed and taking an unconstrained gradient step towards maximizing the objective, and (iii) enforcing the constraint by projecting $\PP$ onto the Fantope (\Cref{def:fantope}), using the algorithm given by \citet{DBLP:conf/nips/VuCLR13}.  And, see \cref{app:optimization} for more details.\looseness=-1

\section{Relation to Interative Nullspace Projection}
\label{sec:inlp}
In this section, we provide an analysis of iterative nullspace projection \citep[INLP;][]{ravfogel-etal-2020-null}, a recent linear concept erasure method that attempts to mitigate bias in pre-trained presentations in a seemingly similar manner to our maximin formulation. 
INLP constructs an orthogonal projection matrix $\PP$ by iteratively training a generalized linear model and projection on the complement of the subspace spanned by the parameter vector.
We give complete pseudocode in \Cref{code:inlp}.
If runs for $\targetdimension$ iterations, INLP returns an orthogonal projection matrix of $D - \targetdimension$.
See \citet{ravfogel-etal-2020-null} for more details.

Given our formulation \Cref{sec:formulation}, we ask the following question: For what pairs of loss and link functions $\loss( \cdot, \cdot )$ and $\link {\cdot}$, does INLP return an exact solution to the objective given in \cref{eq:main-game}?
In the case of linear regression, we give a counter-example that shows that INLP is not optimal in \cref{sec:inlp-lr}.
However, we are able to show that
INLP optimally solves problems with a Rayleigh quotient loss in \cref{sec:inlp-pls}.\looseness=-1

\subsection{Linear Regression}\label{sec:inlp-lr}

In the following proposition,
we show that INLP does not result in the optimal solution given in \Cref{prop:lin}.
While INLP will eventually damage the ability to perform linear regression on the task, it may remove an unnecessarily large number of dimensions.

\begin{proposition}
INLP (\Cref{code:inlp}) applied to linear regression (\Cref{ex:lin}) does not return \Cref{eq:linear-regression-equilibrium}, the orthogonal projection matrix found that is part of the solution found in \Cref{prop:lin} after $\targetdimension=1$ iterations.
\end{proposition}

\begin{proof}
First, stack the training data $\{(\xx{n}, \labely{n})\}_{n=1}^N$ input to INLP into an $N \times D$ matrix $\XX$ and an $N$-dimensional column vector $\yy$.
Then, INLP returns 
\begin{equation}\label{eq:inlp-solution}
\PP_2 = \ID - \frac{(\XX^{\top} \XX)^{-1} \XX^{\top} \yy \yy^{\top}\XX (\XX^{\top} \XX)^{-1}}{\yy^{\top}\XX(\XX^{\top} \XX)^{-\top}  (\XX^{\top} \XX)^{-1} \XX^{\top} \yy},
\end{equation}
applying the standard result about the minimizer of linear regression.
However, in general, \Cref{eq:inlp-solution} is not equal to \Cref{eq:linear-regression-equilibrium}, which proves the result.
\end{proof}

\subsection{Partial Least Squares}\label{sec:inlp-pls}
Interestingly, we find that INLP does recover a solution when applied to partial least squares. 

\begin{proposition}\label{prop:inlp-pls}
INLP (\Cref{code:inlp}) applied to partial least squares (\Cref{ex:pls}) returns the orthogonal projection matrix found that is part of the solution found in \Cref{prop:pls} after $\targetdimension=1$ iterations.
\end{proposition}
\begin{proof}
First, stack the training data $\{(\xx{n}, \labely{n})\}_{n=1}^N$ input to INLP into an $N \times D$ matrix $\XX$ and an $N$-dimensional column vector $\yy$.
Observe that $\XX^{\top}\yy\yy^{\top}\XX$ is real symmetric and of rank 1.
The rest of the proof is effectively a recapitulation of a standard argument of the spectral theorem.\footnote{Note that as an implicit assumption of the proposition, we require $\params$ contains the eigenvectors of $\XX^{\top}\yy\yy^{\top}\XX$.}
Now, consider the first iteration of INLP:
\begin{equation}
\param_1 \in \argmin_{\param \in \params} \frac{\param^{\top}\XX^{\top}\yy\yy^{\top}\XX\param}{||\param||^2}.
\end{equation}
Observe $\param_1$ is an eigenvector associated with the largest eigenvalue \citep{spectral}.
Moreover, $\param_1$ is the \emph{only} non-zero eigenvector. 
INLP returns $\PP_2 = \ID - \frac{\param_1 \param_1^{\top}}{\param_1^{\top} \param_1}$
which is of the form in \Cref{pls-solution2} in \Cref{lemma:rayleigh}. 
This shows the result.\looseness=-1
\end{proof}

\Cref{prop:inlp-pls} is little more than a recapitulation of the fact that Rayleigh quotient problems can be solved by a singular value decomposition, which itself, can be performed iteratively \cite{wold1966estimation}.

\subsection{Logistic Regression}
In \cref{sec:experiments}, we empirically demonstrate that INLP does not return a similar $\PP$ to the orthogonal projection matrix found by INLP.
And, moreover, optimizing \Cref{eq:relaxed-game} generally results in a $\PP$ of lower rank than the one found by INLP that is just as effective at neutralizing the target concept.
Indeed, in all experiments, we were able to identify a 1-dimensional subspace whose removal completely neutralized the concept using \Cref{eq:relaxed-game}, while INLP requires more than one direction.

\begin{figure}[]
    \centering
    \includegraphics[width=0.96\columnwidth]{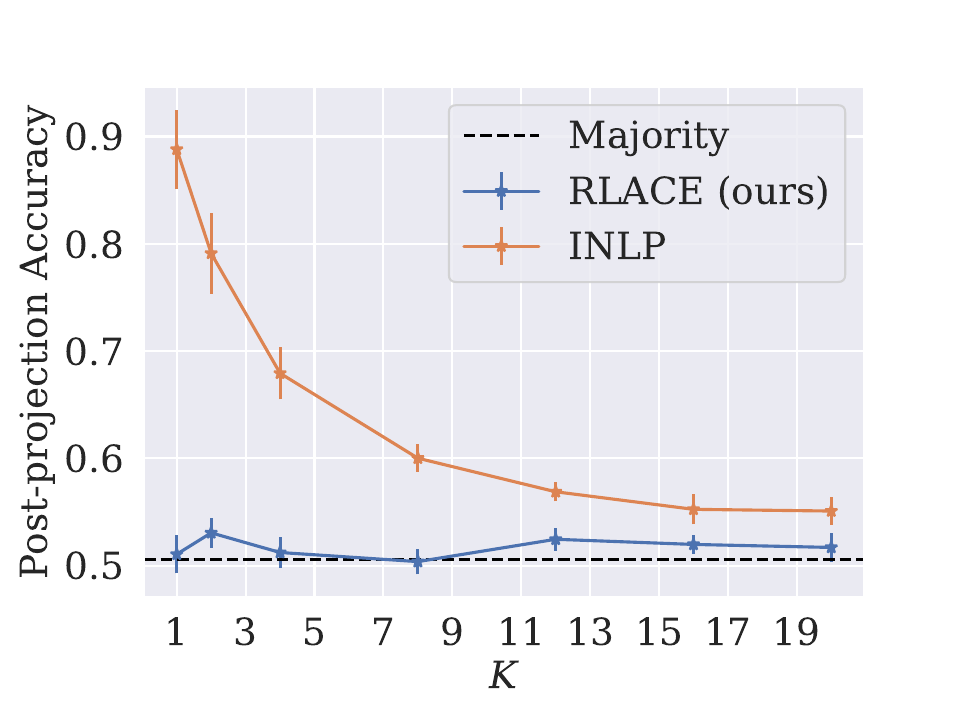}
    \caption{Gender prediction accuracy after bias-removal projection against the dimensionality of the neutralized subspace for INLP and \method,,  on GloVe representations (Experiment \cref{sec:embeddings}). 
    Error bars indicate standard deviation.}
    \label{fig:GloVe}
\end{figure}

\section{Experiments}
\label{sec:experiments}

In this section, we consider mitigating gender associations in static word representations (\S\ref{sec:embeddings}) and increasing fairness in multi-class classification over contextualized representations (\cref{sec:deep-classification}).
Additionally, we qualitatively demonstrate the impact of \method on the input space by linearly removing different concepts from images (\cref{sec:images}).\footnote{In \cref{app:relaxation}, we demonstrate that our method identifies a matrix that is close to a projection matrix, i.e., a vertex of the Fantope.\looseness=-1}

\subsection{Static Word Representations}
\label{sec:embeddings}

We replicate the experimental design of \citet{ravfogel-etal-2020-null} and \citet{gonen-goldberg-2019-lipstick}, which allows the experiment to ascertain the effectiveness of bias mitigation in static word representations. 
Our experiments focus on the static word representations given by the uncased version of the GloVe \cite{DBLP:conf/emnlp/PenningtonSM14}, which \citeposs{ravfogel-etal-2020-null} dataset annotates with a binary label for whether it is male- or female-biased.
See \cref{app:static-setting} for further details about our experimental setting. 
We perform 10 runs of \method and INLP with random initializations and report mean and standard deviations. 

\begin{table}
\centering
\scalebox{0.85}{
\begin{tabular}{lll}
\toprule
    &                \textsf{WEAT's d} $\downarrow$ &                  $p$-value \\
\midrule
  \textsf{Math-art.} \\
  \midrule
 Original &             1.57 &              0.000 \\
      PCA &  1.46 $\pm$ 0.00 &  0.000 $\pm$ 0.000 \\
    RLACE &  0.80 $\pm$ 0.01 &  0.062 $\pm$ 0.002 \\
     INLP &  1.11 $\pm$ 0.10 &  0.015 $\pm$ 0.008 \\
     \hline 
     \textsf{Professions-family.} \\
     \hline
     Original &             1.69 &              0.000 \\
    PCA &  1.11 $\pm$ 0.00 &  0.005 $\pm$ 0.000 \\
    RLACE &  0.79 $\pm$ 0.01 &  0.071 $\pm$ 0.003 \\
     INLP &  1.11 $\pm$ 0.08 &  0.012 $\pm$ 0.007 \\
     \hline
     \textsf{Science-art.} \\
     \hline
      Original &             1.63 &              0.000 \\
      PCA &  1.16 $\pm$ 0.00 &  0.003 $\pm$ 0.000 \\
    RLACE &  0.77 $\pm$ 0.01 &  0.072 $\pm$ 0.004 \\
     INLP &  1.01 $\pm$ 0.15 &  0.028 $\pm$ 0.020 \\
\bottomrule
\end{tabular}
}
\caption{WEAT bias association results.}
\label{tab:weat}
\end{table}

\paragraph{Classification.} 
Before projection, a logistic regressor can recover the gender label of a word representation with near-perfect accuracy. 
However, this accuracy drastically drops after an application of \method for all the different values of $\targetdimension$ examined.
Indeed, post-projection accuracy drops to nearly chance even when we set $\targetdimension=1$; see \cref{fig:GloVe}.
This finding suggests that there exists a 1-dimensional subspace whose removal fully hinders gender classification.\footnote{This result was later proven by \citet{leace}.}
INLP, in contrast, does not reach majority-class accuracy even after the removal of a $20$-dimensional subspace. 
We also examined the PCA-based approach of \citet{bolukbasi2016man}, where the subspace neutralized is defined by the first $\targetdimension$ principal components of the subspace spanned by the difference of the representations for gendered words.\footnote{We used the following pairs of words to compute such a difference:  (\word{woman}, \word{man}), (\word{girl}, \word{boy}), (\word{she}, \word{he}), (\word{mother}, \word{father}), (\word{daughter}, \word{son}), (\word{gal}, \word{guy}), (\word{female}, \word{male}), (\word{her}, \word{his}), (\word{herself}, \word{himself}), (\word{mary}, \word{john}).\looseness=-1}
However, for all $\targetdimension \in \{1, \dots, 10\}$ examined, \citeposs{bolukbasi2016man} method did not significantly influence gender prediction accuracy after applying the orthogonal projection.  
In \citet{ravfogel-etal-2020-null}, it was shown that high-dimensional representation space tends to be (approximately) linearly separable into binary gender by \emph{multiple} different orthogonal linear classifiers.
Our results, surprisingly, show that there is a $1$-dimensional subspace whose removal exhaustively removes the gender concept.  
Importantly, as expected given that we removed a \emph{linear} subspace, non-linear classifiers are still able to recover gender: Both RBF-SVM and a ReLU MLP with 1 hidden layer of size 128 predict gender with above 90\% accuracy.\looseness=-1

\paragraph{Clustering by Gender.}
We now explore how \method influences the geometry of the representation space.
Using \method with $\targetdimension=1$ iterations, we estimate an orthogonal projection matrix that hinders a logistic regression from classifying the representations by their associated gender.
We perform principal components analysis on the GloVe representations before and after applying the orthogonal projection matrix, coloring the points by gender.
As can be seen in \cref{fig:pca}, the original representation clusters by gender, however, this clustering significantly decreases post-projection.
See \cref{app:v-measure} for a quantitative analysis of this effect.\looseness=-1

\paragraph{Word Association Tests.}
\citet{DBLP:journals/corr/IslamBN16} introduce the Word Embedding Association Test (WEAT), a measure for the association of similarity between male- and female-biased words as well as stereotypically gender-biased professions. 
The test examines, for example, whether a group of words denoting STEM professions is more similar, on average, to male names than to female ones. 
We measure the association between stereotypically male and female names and (1) career- and family-related terms, (2) art and mathematics words, and (3) artistic and scientific fields.
We report the test's statistic, \textsf{WEAT's d}, and the associated $p$-values after applying a rank-$1$ projection in \cref{tab:weat}. 
\method is the most effective method.\looseness=-1

\paragraph{Influence on Semantic Content.}
We now test to what extent \method damages the semantic content of the word representations.
We evaluate our word representations using SimLex-999 \citep{DBLP:journals/coling/HillRK15}, a test that measures the quality of the representation space by comparing word similarity in that space to lexical similarity, as judged by human annotators. 
The test is composed of pairs of words, and we calculate the Pearson correlation between the cosine similarity before and after projection, and the similarity score that human annotators gave to each pair. 
Similarly to \citet{ravfogel-etal-2020-null}, we find no significant influence on correlation with human judgments.
Specifically, we observe a Pearson correlation of $0.399$ with the cosine similarity over the original representations, $0.392$ after applying a rank-1 orthogonal projection, estimated by \method, and $0.395$ after 1 iteration of INLP. 
See \cref{app:neighbors} for the neighbors of randomly chosen words before and after \method.

\begin{table*}
\centering
\resizebox{1.98\columnwidth}{!}{%
\begin{tabular}{l|lllll} \toprule
  Setting & Accuracy (gender) $\downarrow$
 &
  Accuracy (Profession) $\uparrow$ &
  $\mathrm{GAP}_{\concept{male},y}^{\mathrm{TPR}, \mathrm{RMS}} \downarrow$ &
  $\sigma_{(\mathrm{GAP}^{\mathrm{TPR}}, \%\concept{women})} \downarrow$ \\ \midrule
BERT-frozen                        & 99.84 & 79.91 & 0.029 & 0.840 \\
BERT-frozen + RLACE (rank 1)      &  52.16 $\pm$ 0.13 & 79.21 $\pm 0.00$   & 0.020 $\pm$ 0.000  & 0.463 $\pm$ 0.005 \\
BERT-frozen + RLACE (rank 50)      &  53.24 $\pm$ 0.73 & 76.73 $\pm$ 1.03 & 0.021 $\pm 0.001$ & 0.426 $\pm$ 0.043  \\
BERT-frozen + INLP (rank 1)        &  99.30 $\pm$ 0.00 & 79.58 $\pm$ 0.01  & 0.028 $\pm 0.000$ & 0.779 $\pm$ 0.014 \\
BERT-frozen + INLP (rank 50)       &  51.95 $\pm$ 0.25 & 71.27 $\pm$ 0.09 & 0.022 $\pm 0.000$ & 0.338 $\pm$ 0.030  \\
\midrule
BERT-finetuned                     & 85.42 $\pm$ 0.05  &  84.71 $\pm$ 0.09 & 0.026 $\pm$ 0.001 & 0.816 $\pm$ 0.005 \\
BERT-finetuned + RLACE (rank 1)    &  53.61 $\pm$ 0.72 &  83.42 $\pm$ 0.10 & 0.022 $\pm$ 0.001 & 0.705 $\pm$ 0.022 &  \\
BERT-finetuned + RLACE (rank 100) &  53.87 $\pm$ 1.32 & 80.93 $\pm$ 1.04 & 0.024 $\pm$ 0.001 & 0.658 $\pm$ 0.030 \\
BERT-finetuned + INLP (rank 1)     &  96.30 $\pm$ 0.63 & 85.41 $\pm$ 0.06  & 0.026 $\pm$ 0.000 & 0.820 $\pm$ 0.007  \\
BERT-finetuned + INLP (rank 100)   & 62.76 $\pm$ 1.31 & 83.74 $\pm$ 0.09  &0.021 $\pm$ 0.001  & 0.579 $\pm$ 0.048 \\
BERT-finetuned-adv (MLP adversary)                          & 98.01 $\pm$ 
1.73 & 83.72 $\pm$ 1.69  & 0.024 $\pm$ 0.003 & 0.707 $\pm$ 0.079 \\
BERT-finetuned-adv (Linear adversary)                          & 99.40 $\pm$ 0.07 & 84.68 $\pm$ 0.16  & 0.026 $\pm$ 0.001 & 0.803 $\pm$ 0.015 \\ \midrule
Majority & 53.52 & 30.0 & - & - &
\\ \bottomrule
\end{tabular}
}
\caption{Results from \cref{sec:deep-classification}.}
\label{tbl:bios-results}
\end{table*}

\begin{figure}[]
    \centering
    \includegraphics[width=0.9\columnwidth]{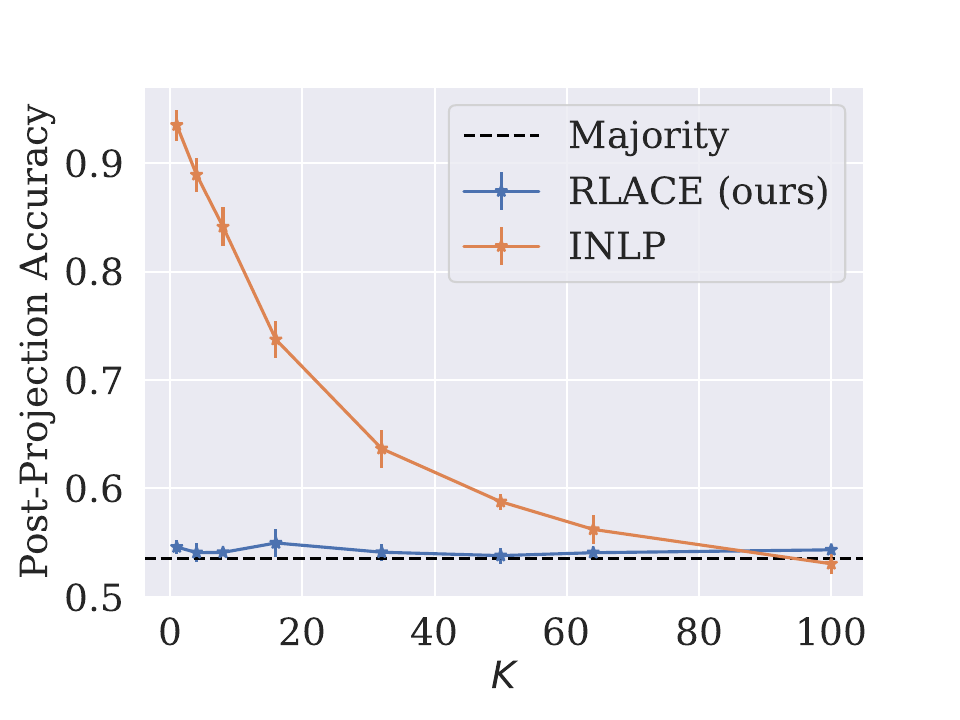}
    \caption{Gender prediction accuracy after bias-removal projection against the dimensionality of the neutralized subspace, for INLP and \method, finetuned BERT representations (Experiment \cref{sec:deep-classification}).}
    \label{fig:bios}
    \vspace{-15pt}
\end{figure}

\subsection{Profession Classification}
\label{sec:deep-classification}
We now evaluate the impact of \method on the fairness of a profession classifier.
We consider \citeposs{DBLP:journals/corr/abs-1901-09451} dataset of short biographies collected from the web, annotated with both binary gender and profession. 
We represent each biography with the \CLS representation in the last layer of BERT \citep{devlin-etal-2019-bert}, apply \method to erase gender from the \CLS representation, and then evaluate the performance of the classifier, after applying the orthogonal projection to the input representations, on the downstream task of profession prediction.

We consider several profession classifiers.
\begin{itemize}
    \item A multiclass logistic regression profession classifier that operates on the frozen representations of pre-trained BERT (\bertafreezed).
    \item A pretrained BERT model finetuned to the profession classification task (\bertafinetuned).
    \item A pretrained BERT model finetuned to the profession classification task, trained adversarially for gender removal with the gradient-reversal layer method of \citet{ganin2015unsupervised} (\bertadv). 
    We consider (1) a linear adversary, and (2) an MLP adversary with $1$ hidden layer of size $300$ and ReLU activations. 
\end{itemize}

We run \method on the representations of \bertafreezed and \bertafinetuned.
We treat \bertadv as a baseline; we report the results of 3 independent runs with random initialization. 
See \cref{app:bios-setting} for more details.\looseness=-1

To measure the bias in the downstream classifier, we follow \citet{DBLP:journals/corr/abs-1901-09451} and apply the TPR-GAP measure, a quantification of the bias in a classifier by considering the difference (GAP) in the true positive rate (TPR) between individuals with different protected attributes, e.g., race or gender.
We use the notation $\mathrm{GAP}_{y, z}^{\mathrm{TPR}}$ to denote the TPR-gap in profession $z$, e.g., \concept{nurse} for a protected group $y$, e.g., \concept{female}.
We also consider $\mathrm{GAP}_{z}^{\mathrm{TPR}, \mathrm{RMS}}$, the RMS of the TPR-gap across all professions for a protected group $y$.
See \cref{app:bios-setting} and \citet{DBLP:journals/corr/abs-1901-09451} for the formal definitions.
To calculate the relation between the bias the model exhibits and the bias in the data, we also compute the correlation between the TPR gap in a given profession and the percentage of women in that profession, denoted as $\sigma_{(\mathrm{GAP}^{\mathrm{TPR}}, \%\concept{women})}$.

The results are summarized in \cref{tbl:bios-results}.  
\method effectively hinders the ability to predict gender from the representations using a rank-$1$ projection because INLP does not completely remove the ability to predict gender from the finetuned model even after $100$ iterations (\cref{fig:bios}). 
Both methods have a moderate negative impact on the task of profession prediction in the finetuned model, while in the frozen model, INLP---but not \method---also significantly damages the task, degrading performance from $79.91$\% to $71.27$\% accuracy. 
Bias, as measured by $\mathrm{GAP}_{y, z}^{\mathrm{TPR}, \mathrm{RMS}}$ is mitigated by both methods to a similar degree, while INLP has some advantage in decreasing $\sigma_{(\mathrm{GAP}^{\mathrm{TPR}}, \%\concept{women})}$. 
For the finetuned model, the decrease in $\sigma_{(\mathrm{GAP}^{\mathrm{TPR}}, \%\concept{women})}$ is moderate.\looseness=-1

Interestingly, the adversarially finetuned models (\bertadv)---both with a linear and an MLP adversary---show somewhat decreased bias as measured by $\mathrm{GAP}_{y, z}^{\mathrm{TPR}, \mathrm{RMS}}$, but do not hinder the ability to predict gender at all.\footnote{During training, the adversaries converged to close-to-random gender prediction accuracy; but this did not generalize to \emph{new} adversaries at test time. 
This phenomenon was observed---albeit to a lesser degree---by \citet{elazar-goldberg-2018-adversarial}.} 
Beyond the effectiveness of \method for selective information removal, we conclude that the connection between the ability to predict gender from the representation, and the TPR-gap metric, is not clear-cut, and requires further study.

\subsection{Erasing Concepts in Image Data}
\label{sec:images}
\begin{figure}[]
    \centering
    \includegraphics[width=0.9\columnwidth]{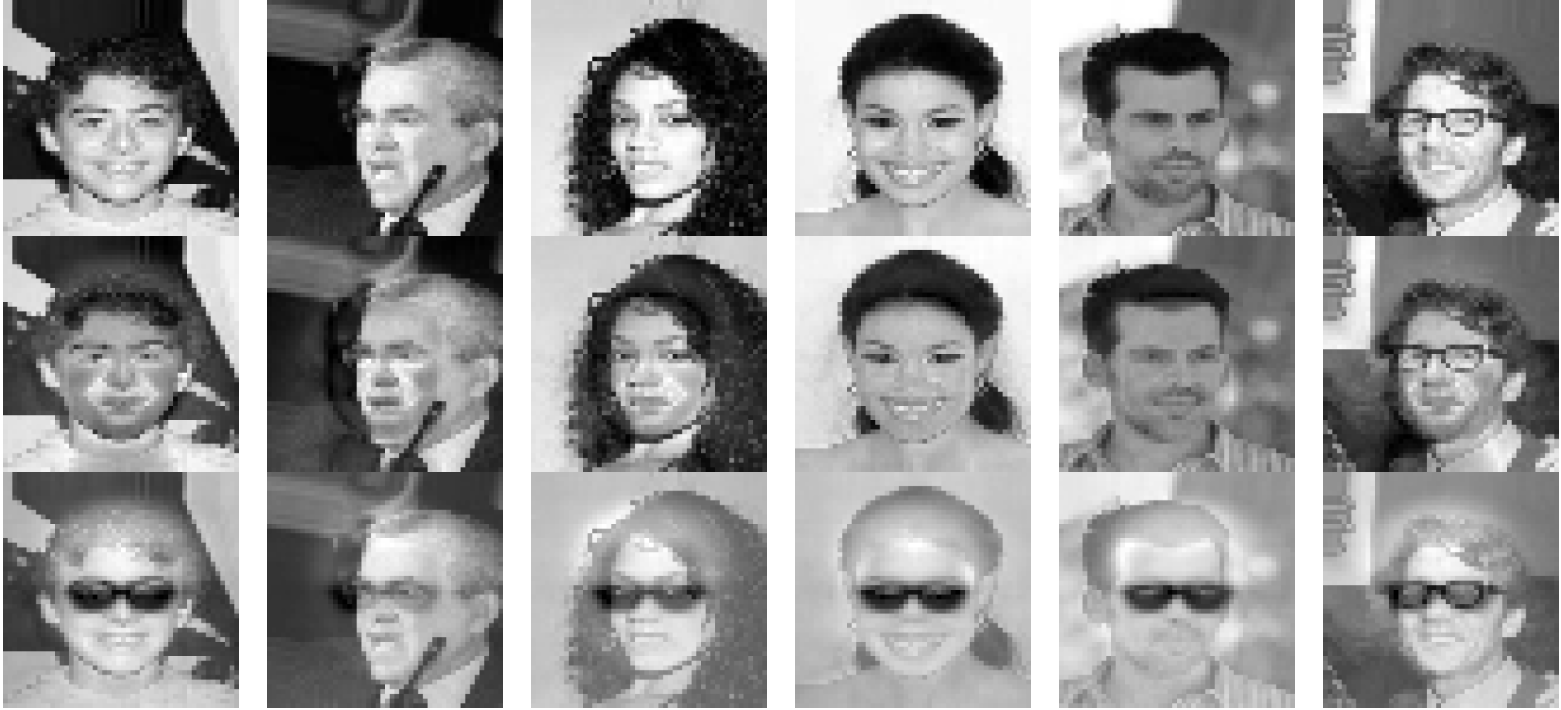}
    \caption{Application of \method to the raw pixels of image data.
 We present images (before and after a rank-$1$ projection) for the concepts \concept{smile} and \concept{glasses}.\looseness=-1}
    \label{fig:faces}
\vspace{-15pt}
\end{figure}

Our empirical focus so far has lain on erasing concepts from textual data. 
We now turn to visual data, which has the advantage of allowing the experiment to be able to clearly inspect the influence of \method on the input. 
To qualitatively assess this effect, we use face images from the CelebsA dataset \cite{DBLP:conf/iccv/YangLLT15}, which is composed of faces annotated with different concepts, such as \concept{sunglasses} and \concept{smile}.
We downscale all data to 50-pixel-by-50-pixel, grey-scale images, flatten them to 2,500-dimensional vectors, and run our method on the pixels and hinder a linear classifier from classifying, for instance, whether a person has sunglasses based on the pixels of their image.\footnote{Modern architectures for computer vision rely on deep models. 
We focus on linear classification in order to better understand the effect of \method.}
We experiment with the following visual concepts: \concept{glasses}, \concept{smile}, \concept{mustache}, \concept{beard}, \concept{bald} and \concept{hat}.

\paragraph{Results.}
See \cref{fig:faces} and \cref{app:faces} for randomly sampled images before and after erasure.
In all cases, a rank-$1$ orthogonal projection matrix, discovered by \method, is enough to remove the ability to classify the images into their concepts.
Indeed, we achieve a classification accuracy of less than 1\% above majority-class accuracy. 
We observe that the intervention changes the images by manipulating the features one would expect to be associated with the concepts of interest, e.g., erasing the concept \concept{sunglasses} results in an image with sunglasses superimposed on the pixels.\footnote{Note that, in contrast to regular style transfer, we prevent classification of the concept. 
At times, (e.g., in the \concept{sunglasses} case), we converge to a solution that always adds the concept.
However, this need not be the case.} 
Because the intervention is constrained to be a projection, it is limited in expressivity, and it is easier to remove features than add new ones.\looseness=-1

\section{Related Work}
Current approaches to concept erasure are predominantly based on adversarial approaches applied at training time \cite{ganin2015unsupervised, edwards2015censoring, chen2018adversarial, xie2017controllable, zhang2018mitigating, wang2021dynamically}. 
However, such methods have proven themselves unstable and were shown by \citet{elazar-goldberg-2018-adversarial} to not completely remove the concept present in the representations.
To the authors' knowledge, the first post-hoc linear concept erasure method was given by \citet{bolukbasi2016man}, who used PCA to identify a gender subspace spanned by a few presupposed gender directions.
Building on the criticism of \citet{gonen-goldberg-2019-lipstick}, several authors have proposed alternative linear formulations \cite{dev2019attenuating, ravfogel-etal-2020-null, dev2020oscar, kaneko2021debiasing}. 
Closest to our work is \citet{sadeghi2019global}, who study a different linear adversarial formulation. 
Their analysis is focused on the special case of linear regression, and they considered a general linear adversary, i.e., one that is not constrained to apply an orthogonal projection matrix, which is more expressive.

Beyond bias mitigation, concept subspaces have been used as an interpretability tool \cite{kim2018interpretability} in the causal analysis of neural networks \cite{elazar2021amnesic,ravfogel-etal-2021-counterfactual} and in the study of the geometry of neural networks' representations \cite{celikkanat2020controlling,gonen2020s, hernandez-andreas-2021-low}. 
Moreover, our linear concept erasure objective is different than past work, which is based on subspace clustering \cite{parsons2004subspace}, because we focus on hindering the ability of a linear classifier to predict the concept from the representations; we do not assume that the data lives in a linear subspace.\looseness=-1

\section{Conclusion}
We have formulated the task of concept erasure from the representation space as a constrained version of a general maximin game. 
In the constrained game, the adversary is limited to a fixed-rank orthogonal projection. 
This constrained formulation allows us to derive closed-form solutions to this problem for certain objectives and to devise a 
more general convex relaxation that works well in practice for others. 
We empirically show that the relaxed optimization recovers a $1$-dimensional subspace whose removal is enough to mitigate linearly encoded concepts. 
As a downside, the method proposed here \emph{only} protects against linear classifiers. 
Effectively removing non-linear information while maintaining the advantages of the constrained, linear approach remains an open challenge. 

\bibliography{icml22}
\bibliographystyle{acl_natbib}

\onecolumn 
\appendix

\section{Ethical Considerations}
\label{app:ethics}
The experiments discussed in this paper revolved around the removal of binary gender information from a representation of text.
First, we would like to acknowledge that gender is a non-binary concept.
Beyond this point, however, erasure of binary gender has real-world applications---in particular, as it relates to fairness in machine learning. 
However, we would like to caution the readers to take the results with a grain of salt and be careful when deploying \method in a practical setting.  
Despite our formal analysis, care should be taken to measure the effectiveness of the approach in the context in which \method is to be deployed, considering, among other things, the exact data to be used, the exact fairness metrics under consideration, and the overall application.
We further urge practitioners not to regard this method as a solution to the problem of bias in representation, but rather as a preliminary research effort towards mitigating certain aspects of the problem.  Unavoidably, we only consider a limited set of datasets in our experiments, and they may not reflect all the subtle and implicit ways in which gender bias may manifest itself.
As such, it is likely that different forms of bias still exist in the representations after applying \method.

\section{Proofs}

\subsection{Proof of \Cref{prop:lin}}\label{app:regression}
\regression*

\begin{proof} 
We lower- and upper-bound the maximin game to prove the result.
\paragraph{Lower Bound.}
First, note that
\begin{equation}
\min_{\param \in \params} ||\yy - \XX \PPzero\param||^2 \leq \max_{\PP \in \projk}  \min_{\param \in \params} ||\yy - \XX \PP\param||^2,
\end{equation}
$\forall \PPzero \in \projk$.
Choose 
\begin{equation}
\PPzero = \ID - \frac{\XX^{\top} \yy \yy^{\top} \XX}{\yy^{\top} \XX \XX^{\top} \yy}.
\end{equation}
Then, we seek to solve the following minimization problem
\begin{equation}\label{eq:optimization}
\min_{\param \in \params} ||\yy - \XX \PPzero\param||^2.
\end{equation}
Because \Cref{eq:optimization} is convex, we only check the first-order optimality condition
\begin{subequations}
    \begin{align}
    \frac{\partial}{\partial \param} ||\yy - \XX \PPzero\param||^2 &= -2 \PPzero \XX^{\top} (\yy - \XX \PPzero\param) \\
     &= \underbrace{-2 \PPzero^{\top} \XX^{\top} \yy}_{=\boldsymbol{0}} +2 \PPzero^{\top} \XX^{\top} \XX \PPzero\param \\
     &= \boldsymbol{0},
    \end{align}
\end{subequations}
which implies we have
\begin{equation}
    \mathbf{0} = \PPzero^{\top} \XX^{\top} \XX \PPzero\param = (\PPzero \XX)^2 \param. 
\end{equation}
Thus, we have a solution at $\param^\star = \boldsymbol{0}$.\footnote{Note that there are other solutions, e.g., we could take $\param = \XX^{\top} \yy$. 
However, because the objective is convex, $\boldsymbol{0}$ is a global minimum. 
}
Plugging $\PPzero$ back into the objective, we arrive at
\begin{equation}
||\yy - \XX \PPzero \boldsymbol{0}||^2 = ||\yy||.
\end{equation}
Thus, we achieve the following lower bound 
\begin{equation}
||\yy|| \leq  \max_{\PP \in \projk} \min_{\param \in \params} ||\yy - \XX \PP\param||^2.
\end{equation}

\paragraph{Upper bound.}
Moreover, because the second player, who seeks to minimize the objective, can always choose $\param^{\star} = \boldsymbol{0}$, we have the following upper bound
\begin{equation}
\max_{\PP \in \projk} \min_{\param \in \params} ||\yy - \XX \PP\param||^2 \leq ||\yy||^2.
\end{equation}

\paragraph{Putting it together.}
Putting the bounds together, we have that $\left(\ID - \frac{\XX^{\top} \yy \yy^{\top} \XX}{\yy^{\top} \XX \XX^{\top} \yy}, \boldsymbol{0}\right)$ is an solution, as desired.
Moreover, we see the value of the 
objective at $\left(\ID - \frac{\XX^{\top} \yy \yy^{\top} \XX}{\yy^{\top} \XX \XX^{\top} \yy}, \boldsymbol{0}\right)$ is $||\yy||^2$.
\end{proof}

\subsection{Proof of \Cref{lemma:rayleigh}}
\label{app:rayleigh}

\rayleigh*

\begin{proof}
First, we manipulate the objective as follows
\begin{subequations}
\begin{align}
    \frac{(\PP\param)^{\top} \AA (\PP\param )}{(\PP\param)^{\top}(\PP\param)} 
    &= \frac{(\PP\param)^{\top} \VV^{\top} \LLambda \VV (\PP\param)}{(\PP\param )^{\top}(\PP\param)}\\
    &= \frac{(\PP\param)^{\top} \VV^{\top} \LLambda \VV (\PP\param)}{(\PP\param)^{\top} \VV^{\top} \VV (\PP\param)}.
\end{align}
\end{subequations}
where we enforce the constraint that $||\PP \param||^2 = 1$.
Note that we are ensured the existence of an eigendecomposition of $\AA$ because we assume $\AA$ symmetric.

\paragraph{Lower Bound.}
To construct a lower bound, choose 
\begin{equation}
    \PPzero = \ID - \sum_{k=1}^{\targetdimension} \eigenvector{d}\eigenvector{d}^{\top}.
\end{equation}
Then, consider
\begin{equation}\label{eq:lower-bound-middle}
      \min_{\param \in \params} \frac{\param^{\top} \PPzero^{\top} \VV^{\top} \LLambda \VV \PPzero \param}{||\PPzero \param||^2} \leq \max_{\PP \in \projk}  \min_{\param \in \params} \frac{\param^{\top} \PP^{\top} \VV^{\top} \LLambda \VV \PP \param}{||\PP \param||^2}.
\end{equation}
The left-hand side of \Cref{eq:lower-bound-middle}, however, is minimized with $\param = \eigenvector{\targetdimension+1}$ and achieves a value of $\eigenvalue{\targetdimension+1}$.
Thus, we have
\begin{equation}
    \eigenvalue{\targetdimension+1} \leq \max_{\PP \in \projk} \min_{\param \in \params}  \frac{\param^{\top} \PP^{\top} \VV^{\top} \LLambda \VV \PP \param}{||\PP \param||^2}.
\end{equation}

\paragraph{Upper Bound.}
We next argue for an upper bound on the objective. 
Choose 
\begin{equation}
 \max_{\PP \in \projk}   \min_{\param \in \params}\frac{(\PP\param)^{\top} \VV^{\top} \LLambda \VV (\PP\param)}{(\PP\param)^{\top} \VV^{\top} \VV (\PP\param)} \leq \max_{\PP \in \projk}  \min_{\param \in \{ \ee{1}, \ldots, \ee{D-k+1}\}}\frac{(\PP\param)^{\top} \VV^{\top} \LLambda \VV (\PP\param)}{(\PP\param)^{\top} \VV^{\top} \VV (\PP\param)}.\label{eq:upper-bound-middle-point}
\end{equation}
We have now reduced the inner continuous minimization problem to a discrete one with $D-\targetdimension+1$ choices.
We can consider each of these choices individually.
Inspection reveals that
\begin{equation}
\max_{\PP \in \projk} \frac{(\PP\ee{j})^{\top} \VV^{\top} \LLambda \VV (\PP\ee{j})}{(\PP\\\ee{j})^{\top} \VV^{\top} \VV (\PP\ee{j})} = \begin{cases}
\eigenvalue{j} & \textbf{if } \ee{j} \in \range(\PP) \\
0 & \textbf{otherwise}.
\end{cases}
\end{equation}
Because we can choose $\PP$ to have a range of dimension at most $D-\targetdimension$, it follows that we should choose it to span $\{\eigenvector{1}, \ldots, \eigenvector{\targetdimension}\}$, the eigenvectors that correspond to the $\targetdimension$ smallest eigenvalues.
This implies that the right-hand side of \Cref{eq:upper-bound-middle-point} has the solution
\begin{subequations}
\begin{align}
    \PP^\star &= \ID - \sum_{k=1}^{\targetdimension} \eigenvector{k}\eigenvector{k}^{\top} \\
    \param^\star &= \eigenvector{\targetdimension+1},
\end{align}
\end{subequations}
and, thus, we arrive at the upper-bound
\begin{equation}
  \max_{\PP \in \projk}   \min_{\param \in \params} \frac{(\PP\param)^{\top} \VV^{\top} \LLambda \VV (\PP\param)}{(\PP\param)^{\top} \VV^{\top} \VV (\PP\param)} \leq \eigenvalue{\targetdimension+1}.
\end{equation}

\paragraph{Putting it Together.}
Given that we have upper and lower bounded the problem with $\eigenvector{\targetdimension+1}$, we conclude that $(\PP^\star, \param^\star)$ is a solution with
\begin{subequations}
\begin{align}
    \PP^\star &= \ID - \sum_{k=1}^{\targetdimension} \eigenvector{k}\eigenvector{\targetdimension}^{\top} \\
    \param^\star &= \eigenvector{\targetdimension
    +1},
\end{align}
\end{subequations}
and, moreover, that the value of the objective if $\eigenvalue{\targetdimension+1}$, as desired.
\end{proof}

\section{Optimizing the Relaxed Objective}
\label{app:optimization}
In this appendix, we describe the optimization of the relaxed objective given in \cref{eq:relaxed-game}.

\subsection{Alternate Optimization with Projected Gradient Descent}
To optimize the relaxed objective \cref{eq:relaxed-game}, we alternate minimization and maximization sticks over $\param$ and $\PP$, respectively. 
On the one hand, we update $\param$ using descent:
\begin{align}
    \param_{t+1} \gets \left(\param_t - \alpha_t \nabla_{\param} \sum_{n=1}^N \loss(\labely{n}, \link{\param^{\top} \PP \xx{n}} \right).
\end{align}
On the other hand, we update $\PP$ with projected gradient ascent:
\begin{align}
    \PP_{t+1} \gets \left(\Pi_{\fantope}(\PP_t + \alpha_t \nabla_{\PP} \sum_{n=1}^N \loss(\labely{n}, \link{\param^{\top} \PP \xx{n}}) \right),
\end{align}
where $\alpha_t$ is the learning rate, and $\Pi_{\fantope}$ is the projection operation onto the Fantope, given in \citet{DBLP:conf/nips/VuCLR13}. 
The following lemma describes how to calculate that projection:
\begin{lemma}[\citet{DBLP:conf/nips/VuCLR13}]
\label{lemma:projection}
Let $\fantope$ be the $\targetdimension$-dimensional fantope; see \cref{def:fantope}, and let $\PP=\sum_{d=1}^D \lambda_d \vv_d \vv_d^{\top}$ be the eigendecomposition of $\PP$ where $\lambda_d$ is $\PP$'s $d$\textsuperscript{th} eigenvalue and $\vv_d$ is its corresponding eigenvector.
The projection of $\PP$ onto the fantope is given by $\Pi_{\fantope}(\PP)=\sum_{d=1}^D \lambda_d^+(\gamma) \cdot \vv_d \vv_d^{\top}$, where $\lambda_d^+(\gamma)= \min\left(\max(\lambda_d - \gamma, 0), 1\right)$ and $\gamma$ satisfies the equation $\sum_{d=1}^D \lambda_d^+(\gamma)=k$.
\label{lem:proj}
\end{lemma}

\Cref{lem:proj} specifies that finding the projection entails performing an eigendecomposition of $\PP$ and finding $\gamma$ that satisfies a set of monotone, piecewise linear equations. 
Becauase we can easily find $\gamma$ where $\sum_{d=1}^D \lambda_d^+(\gamma)> \targetdimension$ and $\gamma$ where $\sum_{d=1}^D \lambda_d^+(\gamma)< \targetdimension$, we can solve the system of equations using bisection. 

Upon termination of the optimization process, we return the closest vertex of the Fantope.
To do so, we perform a spectral decomposition of $\PP$ and return the orthogonal projection matrix $\PP_{\text{final}}$ whose range spans the first $D-\targetdimension$ eigenvectors.
The process is discussed in more detail in \cref{app:optimization}. 

\subsection{Experimental Setup}
\label{app:static-setting}

In this appendix, we describe the experimental setting for those experiments involving static word representations \cref{sec:embeddings}.
We conduct experiments on 300-dimensional uncased GloVe vectors. 
Following \citep{ravfogel-etal-2020-null}, to approximate the gender annotation for the vocabulary, we project all vectors on the $\overrightarrow{\word{he}}-\overrightarrow{\word{she}}$ direction, and take the 7,500 most male-biased and female-biased words.\footnote{Note that $\overrightarrow{\word{he}}$ is the static representation for the word \word{he} and $\overrightarrow{\word{she}}$ is the static representation for the word \word{she}.}
Note that unlike \citep{bolukbasi2016man}, we use the $\overrightarrow{\word{he}}-\overrightarrow{\word{she}}$ direction only to induce approximate gender labels to train \method.

We use the same train--dev--test split as \citet{ravfogel-etal-2020-null}, but discard the gender-neutral words, i.e., we cast the problem as a binary classification. 
We end up with a training set, evaluation set, and test set of sizes 7,350, 3,150, and 4,500, respectively.
We run this procedure for 50,000 iterations with the cross-entropy loss, alternating between an update to the adversary and to the classifier after each iteration. 

The inner optimization problem described in the Fantope projection operation is solved with the bisection method \citep{arfken2011mathematical}.
We train with a simple SGD, with a learning rate of $0.005$, chosen by experimenting with the development set. 
We use a batch size of 128. 
After each 1000 batches, we freeze the adversary, train the classifier to convergence, and record its loss. 
Finally, we return the adversary which yields the \emph{highest} classification loss. 
At test time, we evaluate the ability to predict gender using logistic regression classifiers trained in scikit-learn \citep{scikit-learn}
For the dimensionality of the neutralized subspace, we experiment with the values $\targetdimension=1 \dots 20$ for INLP and \method. 
We perform 10 runs and report mean $\pm$ standard deviation. 

\section{Experimental Setting: Deep Classification}
\label{app:bios-setting}

In this appendix, we describe the experimental setting for the deep classification experiments \cref{sec:classification}.
We use the same train--dev--test split of the biographies dataset considered by \citet{ravfogel-etal-2020-null}, resulting in training, evaluation, and test sets of sizes 255,710, 39,369, and 98,344, respectively.
We run a simple stochastic gradient descent optimization procedure, with a learning rate of $0.005$ and a weight decay of $1e^{-4}$, chosen by experimenting with the development set. 
We consider a batch size of 256, and, again, choose the adversary which yields the highest classification loss. 
As the dimensionality of the bias subspace, we run both \method and INLP with $\targetdimension=1 \dots 50$ on \bertafreezed and $\targetdimension=1 \dots 100$ on \bertafinetuned.
We perform 3 runs of the entire experimental pipeline (classifier training, applying INLP and \method) and report mean $\pm$ the standard deviation. 

\paragraph{Classifier training.}
We experiment with several profession classifiers, as detailed in \cref{sec:classification}. 
For \bertafreezed, we use the HuggingFace implementation \cite{wolf2019huggingface}. 
For \bertafinetuned, we finetune the pre-trained BERT on the profession classification task, using an SGD optimizer with a learning rate of $0.0005$, weight decay of $1e^{-6}$ and momentum of $0.9$.
We train for $70,000$ batches of size $10$ and choose the model that achieved the lowest loss on the development set. 
For \bertadv, we perform the same training procedure but add an additional classification head which is trained to predict gender, and whose gradient is reversed \cite{ganin2015unsupervised}. 
This procedure creates an encoder that generates hidden representations which are predictive of the professions but are not predictive of gender.
The adversary always converged to a low gender classification accuracy (below $55\%$), which is commonly interpreted as the success of the removal process.

\paragraph{Fairness Measure: TPR-GAP.}
We informally describe the fairness measures used in \cref{sec:deep-classification}.
The TPR-GAP is tightly related to the notion of fairness by equal opportunity \citep{equal-opportunities+odds}: a fair binary classifier is expected to show similar success in predicting the task label $\yy$ for two populations when conditioned on the true class. 
We refer the reader to \citet{equal-opportunities+odds} for more information.

\section{$V$-Measure}
\label{app:v-measure}
\begin{figure}
    \centering
    \includegraphics[width=0.5\columnwidth]{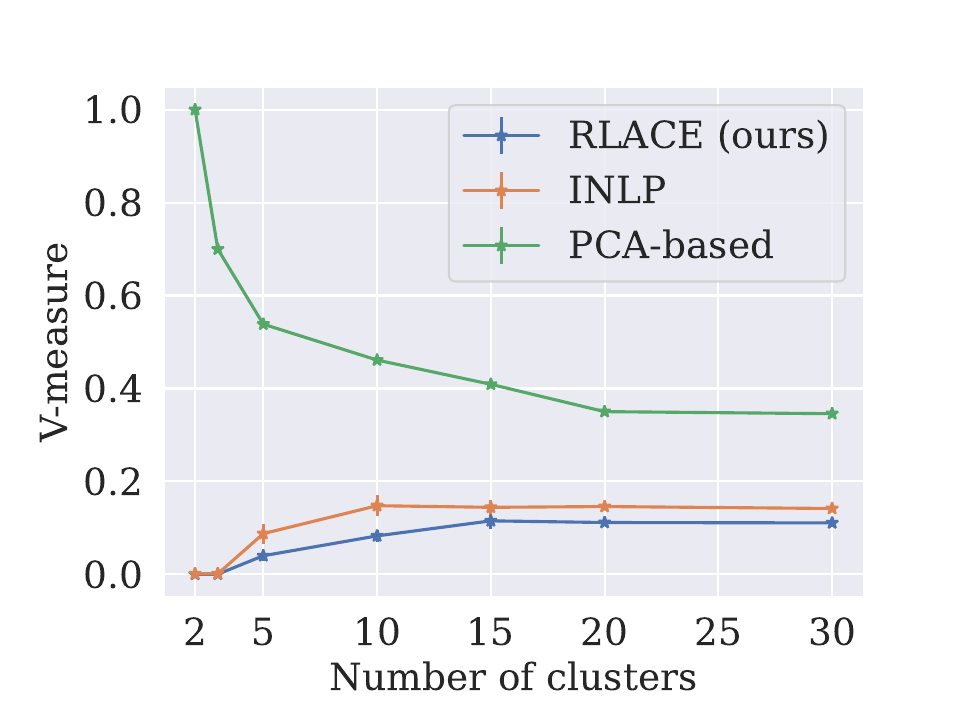}
    \caption{V-measure between gender labels and cluster identity, for different numbers of clusters on the $x$-axis (lower values are better). 
    Error bars are standard deviations from 10 random runs.}
    \label{fig:v-measure}
\end{figure}

To quantify the effect of our intervention on the GloVe representation space in \cref{sec:embeddings}, we perform $k$-means clustering with different values of $k$, and use $V$-measure \cite{DBLP:conf/emnlp/RosenbergH07} to quantify the association between cluster identity and the gender labels, after a projection that removes rank-$1$ subspace. The results are presented in \cref{fig:v-measure}. 
$V$-measure for the original representations is 1.0, indicating strong alignment between cluster identity and gender label. 
The score drastically drops after a rank-$1$ relaxed projection, while INLP projection and the PCA-based method of \cite{bolukbasi2016man} have a smaller effect.

\section{Influence on Neighbors in Embedding Space}
\label{app:neighbors}

In \cref{sec:embeddings}, we showed that the SimLex999 evaluation does not indicate that our intervention damages the general semantics encoded in the GloVe embedding space.
To qualitatively demonstrate this, we provide in \cref{tab:neighbors} the closest-neighbors to 15 randomly sampled words from the vocabulary, before and after our intervention.
\begin{table}[t]
\centering
\begin{tabular}{lll}
\toprule
          Word &                         Neighbors before &                          Neighbors after \\
\midrule
         \word{ocean} &                    \word{waters}, \word{atlantic}, \word{sea} &                    \word{waters}, \word{atlantic}, \word{sea} \\
        \word{museum} &                \word{heritage}, \word{art}, \word{exhibition} &                \word{heritage}, \word{art}, \word{exhibition} \\
           \word{lol} &                             \word{:p}, \word{:d}, \word{haha} &                             \word{:p}, \word{:d}, \word{haha} \\
        \word{twenty} &                       \word{five}, \word{ten}, \word{hundred} &                       \word{five}, \word{ten}, \word{hundred} \\
        \word{sample} &                      \word{free}, \word{test}, \word{samples} &                   \word{example}, \word{test}, \word{samples} \\
         \word{storm} &                      \word{weather}, \word{wind}, \word{rain} &                      \word{weather}, \word{wind}, \word{rain} \\
         \word{state} &                 \word{ohio}, \word{government}, \word{states} &                 \word{ohio}, \word{california} \word{states} \\
    \word{electrical} &        \word{electricity}, \word{mechanical}, \word{electric} &        \word{electricity}, \word{mechanical}, \word{electric} \\
        \word{papers} &                     \word{essay}, \word{essays}, \word{paper} &                     \word{essay}, \word{essays}, \word{paper} \\
 \word{contributions} &  \word{participation}, \word{contribute}, \word{contribution} &  \word{participation}, \word{contribute}, \word{contribution} \\
           \word{lab} &            \word{research}, \word{science}, \word{laboratory} &            \word{research}, \word{science}, \word{laboratory} \\
          \word{joke} &                     \word{laugh}, \word{stupid}, \word{funny} &                     \word{laugh}, \word{stupid}, \word{funny} \\
          \word{hear} &                      \word{tell}, \word{listen}, \word{heard} &                      \word{tell}, \word{listen}, \word{heard} \\
        \word{detail} &           \word{description}, \word{detailed}, \word{details} &           \word{description}, \word{detailed}, \word{details} \\
       \word{extreme} &                \word{hardcore}, \word{severe}, \word{intense} &                \word{hardcore}, \word{severe}, \word{intense} \\
\bottomrule
\end{tabular}
\caption{Neighbors to random words in GloVe space before and other rank-1 \method projection.}
\label{tab:neighbors}
\end{table}

\subsection{Additional results on the CelebsA dataset}
\label{app:faces}
We present here randomly sampled outputs for the 6 concepts we experimented with the following concepts: \concept{glasses}, \concept{smile}, \concept{mustache}, \concept{beard}, \concept{bald}, and \concept{hat};
see the experimental designs in \cref{sec:images}.

\begin{figure}[h]
    \centering
    \includegraphics[width=1\columnwidth]{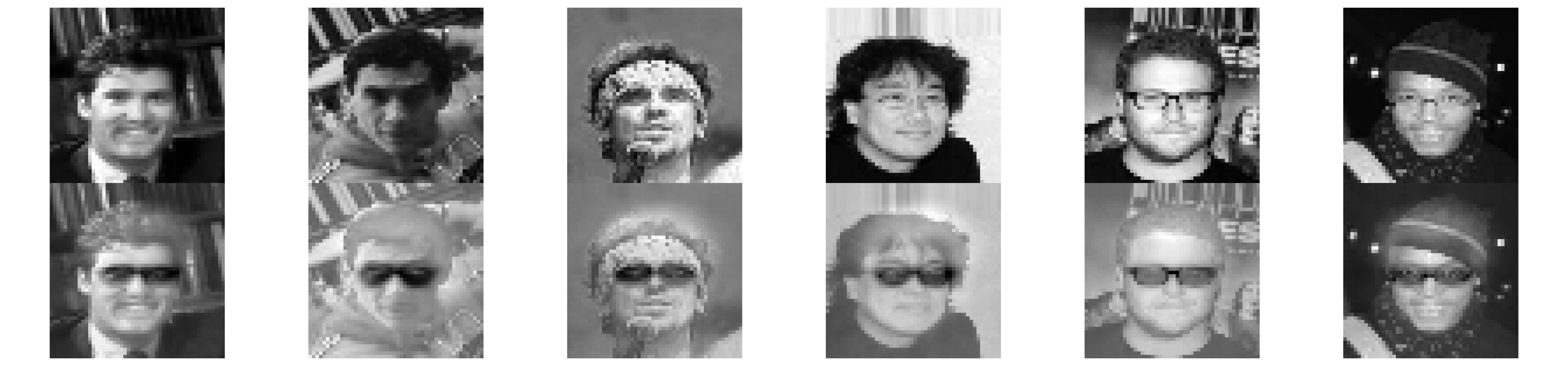}
    \caption{\concept{glasses}}
\end{figure}
\begin{figure}[h]
    \centering
    \includegraphics[width=1\columnwidth]{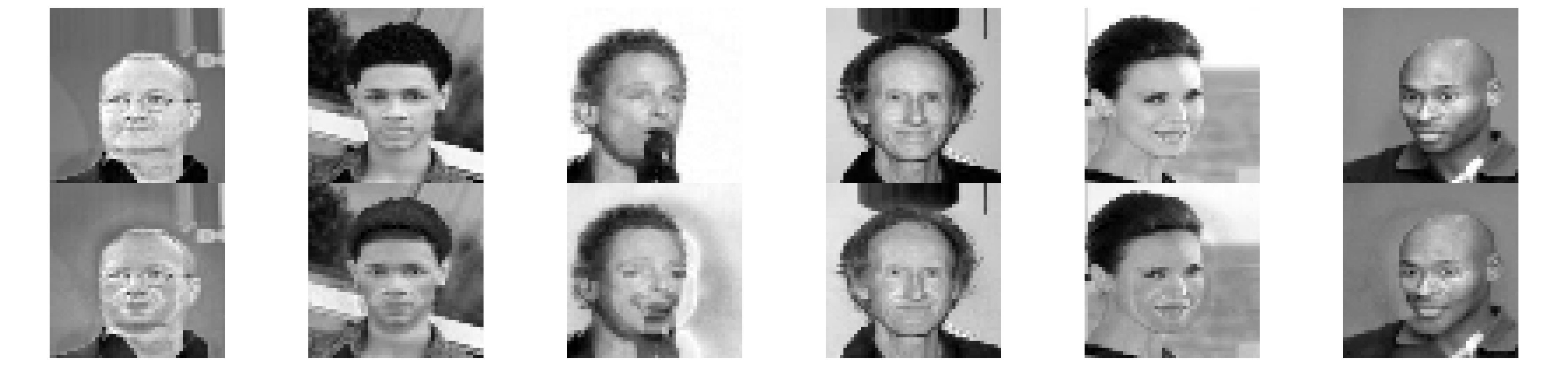}
    \caption{\concept{smile}}
\end{figure}
\begin{figure}[h]
    \centering
    \includegraphics[width=1\columnwidth]{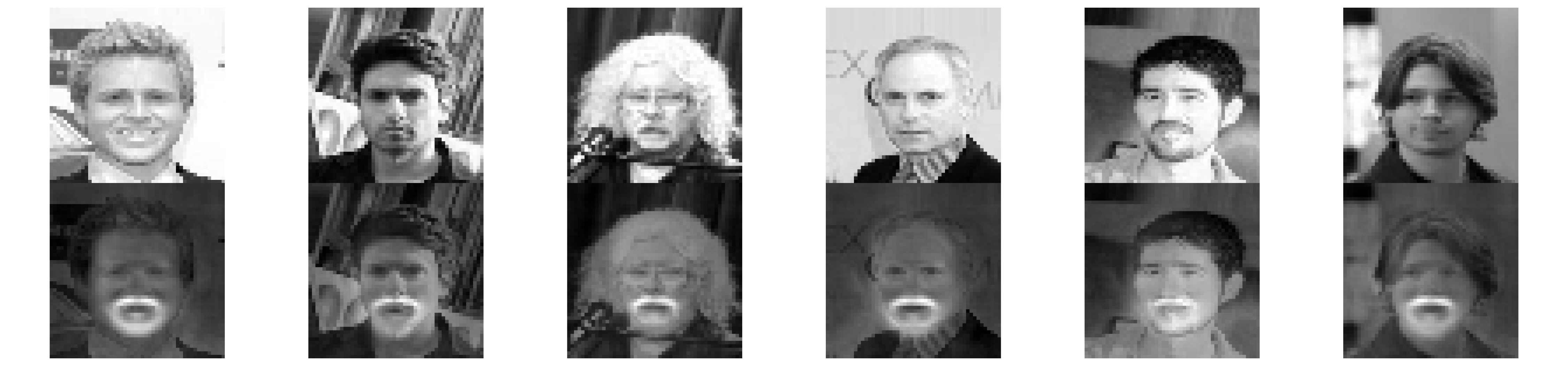}
    \caption{\concept{mustache}}
\end{figure}
\begin{figure}[h]
    \centering
    \includegraphics[width=1\columnwidth]{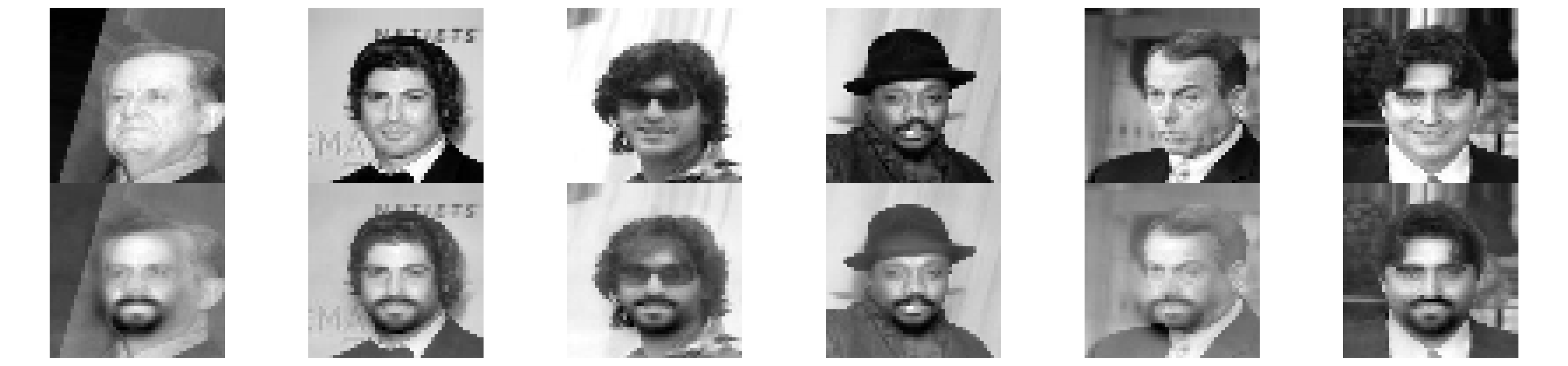}
    \caption{\concept{beard}}
\end{figure}
\begin{figure}[h]
    \centering
    \includegraphics[width=1\columnwidth]{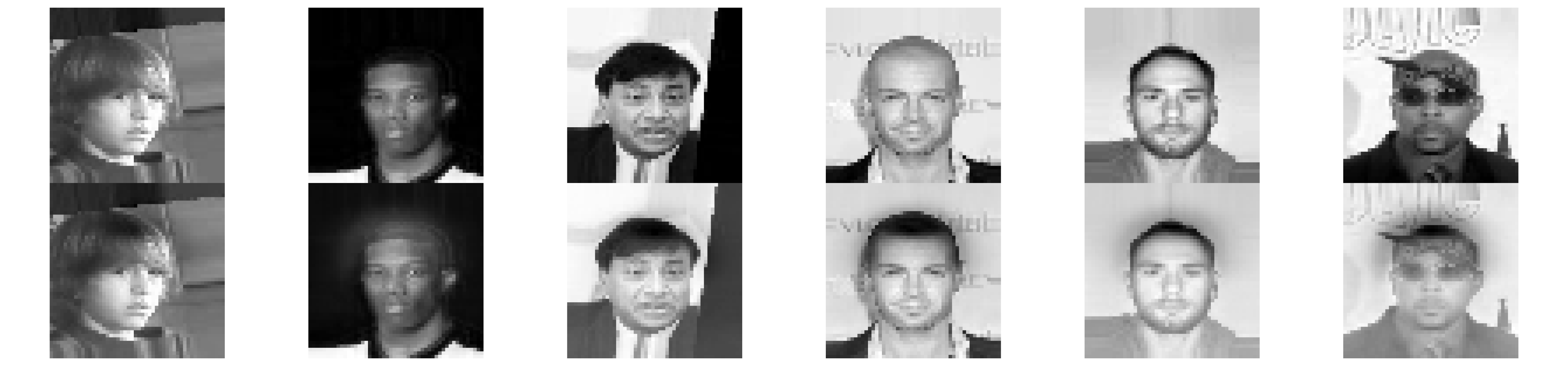}
    \caption{\concept{bald}}
\end{figure}
\begin{figure}[h]
    \centering
    \includegraphics[width=1\columnwidth]{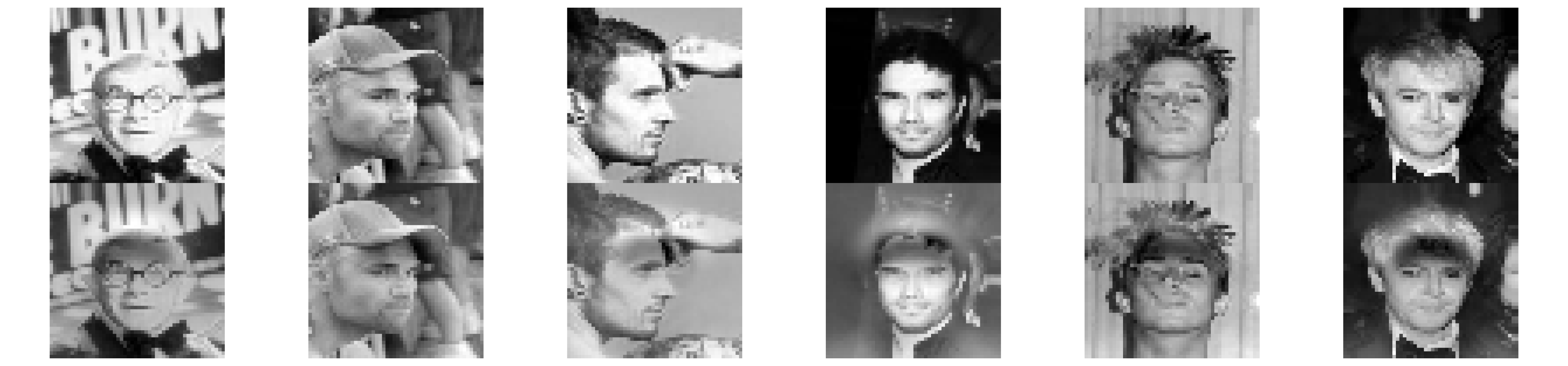}
    \caption{\concept{hat}}
\end{figure}

\subsection{Relaxation Quality}\label{app:relaxation}

We now investigate to what extent the optimization of the relaxed objective \cref{eq:relaxed-game} results in a matrix $\PP$ that is a valid rank-$\targetdimension$ orthogonal projection matrix.
\begin{figure}
    \centering
    \includegraphics[width=0.4\columnwidth]{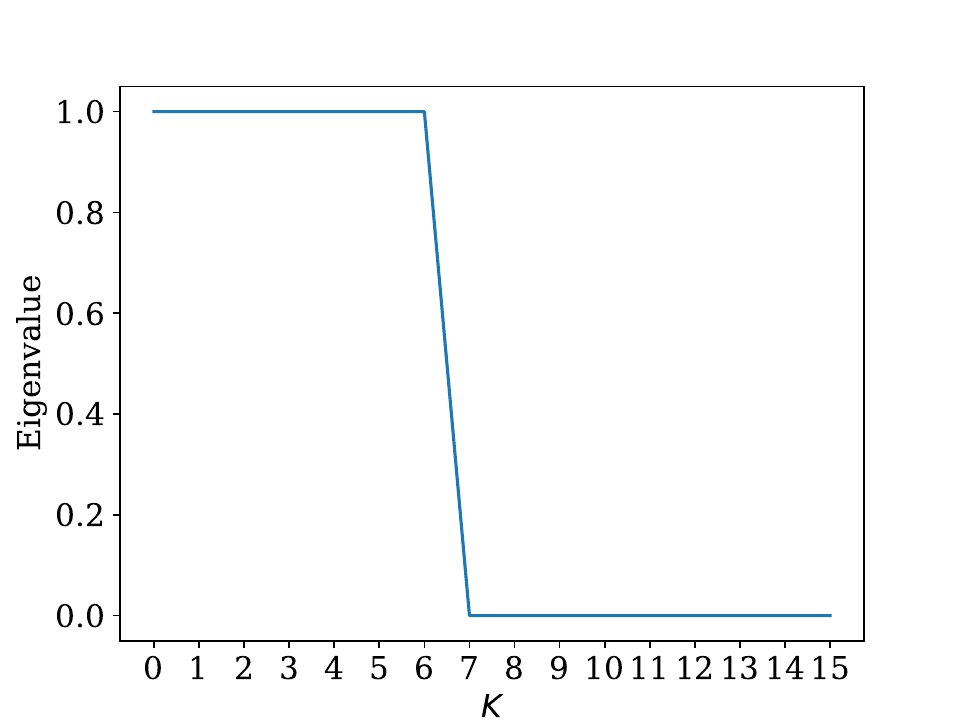}
    \caption{The spectrum of the approximate solution to the relaxed optimization problem \Cref{eq:relaxed-game} with  $\targetdimension=7$.}
    \label{fig:eigenvalues}
\end{figure}
Recall that orthogonal projection matrix have eigenvalues that are either $0$ or $1$, and, accordingly, their sum is the rank of the matrix. 
In \cref{fig:eigenvalues}, we present the eigenvalues spectrum of $\PP$ after optimization with $\targetdimension=6$ on the static word representation dataset (\cref{sec:embeddings}). We find that the top $6$ eigenvalues are indeed close to 1, and the rest are close to 0---suggesting the approximation is tight: the resulting matrix is close to a valid rank-$\targetdimension$ orthogonal projection matrix.\looseness=-1

\end{document}